\pdfoutput=1

\documentclass[letterpaper]{article}

\newcommand{\cv}[1]{}
\newcommand{\av}[1]{#1}

\usepackage{comment}

\excludecomment{cvblock}\includecomment{avblock}

\usepackage[T1]{fontenc}
\usepackage[utf8]{inputenc}

\usepackage{geometry}
\usepackage{newtxtext}
\usepackage{helvet}
\usepackage{courier}
\makeatletter
\let\@dblfloat\@float
\let\end@dblfloat\end@float
\makeatother
\usepackage[tableposition=top]{caption}
\date{}
\newenvironment{IEEEkeywords}%
    {\vspace{0.6\baselineskip}\noindent\small\textbf{Keywords---}}%
    {\par\vspace{0.4\baselineskip}}

\usepackage{cite}
\usepackage{amsmath,amssymb,amsfonts}
\usepackage{graphicx}
\usepackage{url}
\usepackage{xurl}
\usepackage{hyperref}
\hypersetup{
  hidelinks
}
\hypersetup{%
  pdftitle={Improving Constraint Models with LLM Agents},%
  pdfauthor={Florentina Voboril, Stefan Szeider},%
  pdfkeywords={Large Language Models, ReAct Agent, Constraint Model Optimization, CPMpy}}
\usepackage{microtype}
\usepackage{listings}
\usepackage{framed}
\usepackage{booktabs}
\usepackage{threeparttable}
\makeatletter
\def\TPT@opt@flushleft{%
\def\TPTnoteSettings{\labelsep\z@ \leftmargin\z@ \labelwidth\z@}%
\def\TPTnoteLabel##1{\tnote{##1}\hfil}%
\rightskip\z@skip \leftskip\z@skip}
\makeatother
\usepackage{xcolor,colortbl}
\usepackage{tikz}

\usepackage{pgfplots}
\pgfplotsset{compat=1.18}
\usetikzlibrary{positioning,arrows.meta}

\definecolor{better}{HTML}{A5D6A7}
\definecolor{equal}{HTML}{FFE082}
\definecolor{worse}{HTML}{EF9A9A}
\definecolor{nosolution}{HTML}{B0BEC5}
\definecolor{solvererror}{HTML}{90A4AE}
\definecolor{shadecolor}{gray}{0.9}

\begin{document}

\title{Improving Constraint Models with LLM Agents%
\av{\thanks{This is the authors' version of a paper to appear in the proceedings of the 38th IEEE International Conference on Tools with Artificial Intelligence (ICTAI 2026).}}}

\cv{
\author{\IEEEauthorblockN{Florentina Voboril \qquad Stefan Szeider}
\IEEEauthorblockA{
Algorithms and Complexity Group\\
TU Wien, Vienna, Austria\\
\texttt{\{fvoboril,sz\}@ac.tuwien.ac.at}
}}
}
\av{
\author{
    Florentina Voboril \qquad Stefan Szeider\\[4pt]
    \small Algorithms and Complexity Group\\[-2pt]
    \small TU Wien, Vienna, Austria\\[-2pt]
    \small \texttt{\{fvoboril,sz\}@ac.tuwien.ac.at}
}
}

\maketitle
\av{\thispagestyle{empty}}

\begin{abstract}
The runtime of Constraint Programming (CP) solvers is highly sensitive to modeling choices, such as symmetry breaking, implied constraints, global constraints, constraint reformulation, and variable representation. Improving these constraint models has traditionally required human expertise, and existing automated reformulation systems are restricted to a predefined library of hand-crafted transformation rules. We introduce an agentic framework that instead reformulates a constraint model from an open-ended space and establishes correctness empirically rather than by construction: a Large Language Model (LLM) agent, given a model and three training instances, proposes alternative formulations, validates each by injecting its solution back into the original model, and diagnoses and repairs failures, returning the best variant it finds in about six to ten minutes. The models are expressed in the CPMpy modeling library. We evaluate the approach
on fifteen combinatorial optimization problems, including three newly designed problems, with ten independent runs per problem and ten test instances each. The generated models outperform the originals in 1028 of 1500 cases (68.5\%). A non-agentic baseline that samples the same number of candidate models and selects among them with the same validation tools reaches a comparable improvement rate. These results demonstrate that autonomous agentic methods can support the improvement of constraint models.
\end{abstract}


\section{Introduction}

Constraint Programming (CP) is a powerful paradigm for solving combinatorial optimization problems. The solving performance strongly depends on how the problem is modeled. Applying optimization techniques such as adding symmetry-breaking constraints, which eliminate redundant equivalent solutions~\cite{handbook_symmetry}, implied constraints, which strengthen reasoning without changing the solution set~\cite{handbook_implied}, using high-level global constraints with specialized propagation algorithms~\cite{handbook_global_constraints}, reformulating the model, or using alternative variable representations often reduces the search space and speeds up solving time~\cite{cp_handbook_2006}. However, identifying such techniques typically requires substantial expertise and experience in constraint modeling.

There have been previous approaches for automatic reformulation and model improvement in CP. For example, Conjure automatically produces constraint models from an abstract problem specification written in Essence~\cite{akgun_conjure_2022} and Savile Row automatically improves constraint models through predefined reformulation rules and simplifications~\cite{nightingale_savilerow_2017}. There have also been approaches for automatically detecting symmetries~\cite{automatic_symmetry_detection} and generating streamliner portfolios~\cite{spracklen_automated_2023}. Beyond model reformulations, there are also approaches for automated algorithm selection and automated solver tuning. SATzilla~\cite{satzilla_2008} automatically selects promising solvers for SAT on a per-instance basis, whereas approaches like AutoFolio~\cite{lindauer_autofolio_2015} generalize algorithm selection to a broader range of optimization domains. SMAC~\cite{hutter_smac_2011} automatically configures solver parameters to improve performance on benchmark distributions. Unlike these approaches, which select algorithms, tune solver parameters for a fixed constraint model, or apply predefined transformations, our framework directly modifies the model itself: the agent autonomously decides which modeling optimizations to attempt and validates them empirically.

Recent progress in Large Language Models (LLMs) has enabled new approaches at the intersection between CP and generative AI. LLMs have recently been used for active constraint acquisition~\cite{mechqrane_active_2026} and dealing with hard constraints in LLM-generated sentences~\cite{regin_combining_2024}. Another promising direction is to generate CP models from natural language descriptions~\cite{michailidis_constraint_2024, DBLP:conf/ecai/MichailidisTG25, song_llms_2026, szeider_cp-agent_2026}, and it was studied how small changes in the natural language problem description affect generated CP models~\cite{pellegrino_when_2025}. However, little work has examined LLMs' ability to improve existing models. A first step in this direction has been the use of LLMs to generate streamlining constraints for constraint satisfaction problems~\cite{voboril_generating_2025, voboril_streamllm_2025} and optimization problems~\cite{voboril_balancing_2025}.

In this work, we introduce a novel agentic approach that autonomously improves existing constraint models. It goes beyond adding symmetry-breaking or streamlining constraints, as it reformulates the entire model, and beyond classical reformulation systems such as Savile Row~\cite{nightingale_savilerow_2017}, as it draws on an open-ended space of reformulations rather than a fixed library of hand-coded rules. We express the models in CPMpy~\cite{cpmpy}, an open-source Python library for solving CP problems: one can describe problems at a high level and solve them using one of the supported solvers. Recent research has shown that CPMpy is particularly well-suited for LLMs, as they have seen extensive Python code during training~\cite{DBLP:conf/ecai/MichailidisTG25}. We implement our approach using an agent that follows a Reason and Act (ReAct) paradigm~\cite{yao_react_2022}. In ReAct-style agents, an LLM alternates between reasoning about the task and executing the provided \textit{tools}. Each tool behaves like a function with defined input parameters and return values following a JSON schema. Compared to a hard-coded workflow, the agent decides which tool to call in which order.

\textit{Problem Statement.}
We consider the following setting: the user provides a CPMpy model and three small training instances. We want to automatically obtain a modified model whose solutions are feasible under the original model, while improving performance. We focus exclusively on optimization problems, where performance is measured by objective quality within a fixed time limit. The training instances may be used to guide the improvement process, while separate test instances are used only for evaluation.

Our approach to this problem is to utilize an agent that analyzes the original model and iteratively generates alternative model variants by applying potential optimization strategies. It empirically validates them using an automated procedure that checks both correctness and solver performance. In the end, it selects the model it considers best-performing and returns it. We refer to this model as the \textit{proposed model}. The main difficulty is reliability: unlike a curated rule set, an LLM proposes reformulations from an unbounded space and does so unreliably, since some candidates are buggy or silently change the problem. Ensuring that a proposed model is both correct and faster is therefore the central challenge, which the validation-and-repair loop is designed to meet.

To evaluate our approach, we apply it to fifteen combinatorial optimization problems, including several well-known benchmark problems and three novel problems designed specifically for this study. Across these problems, the agent proposes the final model in about six to ten minutes. These models outperform the original models in 1028 of 1500 cases (68.5\%) and yield a mean relative objective change of 14.7\%, demonstrating that LLM agents can successfully identify useful modeling improvements even with only a small number of training instances.

\textit{Main Contributions.}
First, we frame and demonstrate a new point in the design space of automatic constraint-model improvement: reformulation from an \emph{open-ended} space, with correctness established \emph{empirically} rather than through a fixed library of hand-crafted rules. Classical systems such as Savile Row~\cite{nightingale_savilerow_2017} can only apply transformations that have been pre-encoded and proved sound; our agent instead proposes arbitrary reformulations, validates each one empirically, and diagnoses and repairs failures in a Reason-and-Act loop. To our knowledge, this is the first such framework for improving existing constraint models. Second, we provide an empirical validation procedure that checks candidate solutions against the original model by injecting them back into it. Third, on fifteen combinatorial optimization problems the agent's proposed models outperform the originals in 1028 of 1500 cases (68.5\%) and yield a mean relative objective change of 14.7\% (Table~\ref{tab:main_results}), and in none of these comparisons did a proposed model return a solution that violates the original model. Fourth, a comparison of nine LLMs from six providers on four of the problems shows that the approach is not tied to a particular model (Table~\ref{tab:llm_comparison}).
Finally, to assess how much of these gains requires the agentic loop, we compare the agent against non-agentic baselines that sample candidate models independently and reuse the same validation and selection tools; with the same number of candidate models, they reach a comparable improvement rate (Table~\ref{tab:bestofk}).

\section{Agentic Approach}

\begin{figure}[t]
\centering
\begin{tikzpicture}[
  >={Stealth[round]},
  node distance=3.5mm,
  box/.style={draw, rounded corners, align=center, font=\footnotesize,
              inner sep=3pt, text width=5.5cm},
  dec/.style={draw, rounded corners, align=center, font=\footnotesize,
              inner sep=3pt, text width=5.5cm, fill=gray!10}]
  \node[box] (in) {\textbf{Input:} original CPMpy model + 3 training instances};
  \node[box, below=of in] (gen) {\textbf{Generate} an alternative model\\
        {\scriptsize (analyze the model; pick a technique: symmetry breaking, global / implied constraints, reformulation, \ldots)}};
  \node[box, below=of gen] (val) {\textbf{Validate} on the training instances:\\
        solve the alternative model, inject its solution into the original };
  \node[box, below=of val] (cmp) {\textbf{Compare} against the original\\
        {\scriptsize $\Rightarrow$ better / equal / worse}};
  \node[dec, below=of cmp] (dec) {five \emph{correct} models generated?};
  \node[box, below=of dec] (out) {\textbf{Select} and return the proposed model};
  \draw[->] (in) -- (gen);
  \draw[->] (gen) -- (val);
  \draw[->] (val) -- node[right,font=\scriptsize] {correct} (cmp);
  \draw[->] (cmp) -- (dec);
  \draw[->] (val.west) -- ++(-0.3,0)
        node[left,font=\scriptsize,align=right] {invalid /\\ no\_solution /\\ solver\_error:\\ diagnose\\ \& repair}
        |- (gen.west);
  \draw[->] (dec) -- node[right,font=\scriptsize] {yes} (out);
  \draw[->] (dec.east) -- ++(0.3,0)
        node[right,font=\scriptsize,align=left] {no}
        |- (gen.east);
\end{tikzpicture}
\caption{Overview of the agent's Reason-and-Act loop. Through its tools, the LLM agent repeatedly proposes an alternative model and validates it against the original; on failure it diagnoses and repairs, iterating until five correct models have been produced, then returns the one it considers best.}
\label{fig:overview}
\end{figure}
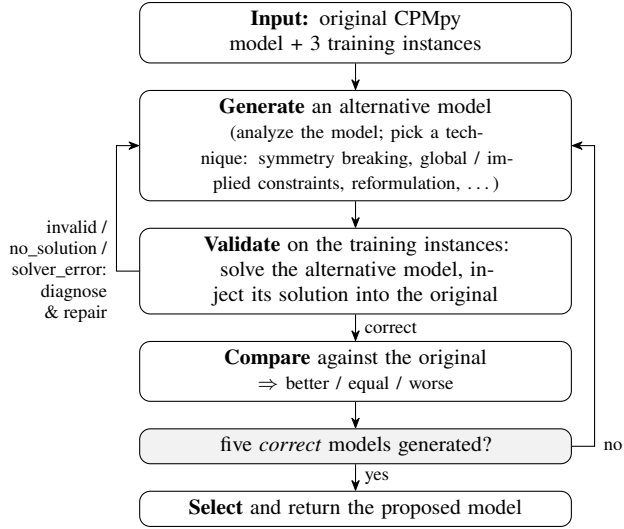

In our approach, we run an open-source agentic Python coder~\cite{szeider_cp-agent_2026} to automatically generate an improved CPMpy model. We decided to use this lightweight agent instead of full-featured coding agents, since the framework provides a high level of transparency, full control over the prompts, better reproducibility, supports various underlying LLMs, and is specifically designed for generating single-script Python code. Since our tasks focus on model generation rather than large-scale software engineering, more complex coding agents would be unnecessary. Figure~\ref{fig:overview} gives an overview of the resulting loop.

\subsection{The Agent's Tools}
\av{\enlargethispage*{6mm}}

The agent has access to the three tools that come with the coder and are not specific to this work: a Python execution tool, which executes Python code in a persistent IPython kernel, a code saving tool, which saves final verified Python code to a file, and a to-do tool, which creates a set of to-do items. The to-do tool forces the agent to plan before coding, track progress step-by-step, and ensure it does not skip important steps.

\subsection{Validation Function}
\label{validation}

The validation function empirically evaluates alternative models for two aspects: a model must be \textit{semantically correct} (i.e., every solution it produces must also be feasible under the original model), and its \textit{performance} is compared against the original model. The function, which the agent calls through the Python execution tool, applies the following three steps to every training instance. Every solve is given a 20\,s time limit.

\textbf{Step 1: Run the alternative model.} This step acts as an early filter. If the alternative model does not follow formatting requirements or has a syntax error, the outcome is \textit{solver\_error}. Likewise, if the model cannot find any solution within the timeout, the outcome is labeled \textit{no\_solution}. Otherwise, a candidate solution $x'$ is obtained and passed to Step~2.

\textbf{Step 2: Run the original model with the injected solution $x'$.} This step verifies the correctness of the new solution by injecting $x'$ back into the original model. All output decision variables of the original model are fixed to the corresponding values in $x'$ via equality constraints, and the augmented model is solved. If the original model is feasible under these constraints, $x'$ satisfies all original problem constraints and is considered correct. Otherwise, the alternative model has produced an invalid solution and is not semantically equivalent to the original model. In this case, the outcome is labeled \textit{invalid} and Step~3 is skipped.

\textbf{Step 3: Compare to original model.} The original model is solved without any injected constraints to obtain a reference objective value. It is computed once per instance and cached for subsequent evaluations. The function then reports whether the objective value obtained by the alternative model in Step~1 is \textit{better} than, \textit{equal} to, or \textit{worse} than this reference.

\subsection{Procedure}

We provide the agent with the original CPMpy model, three small training instances, access to the \texttt{cpmpy} and \texttt{numpy} packages and the validation function, and a project prompt. The agent is able to call the provided tools in any order. However, the prompt gives the agent a guideline on how to proceed:

\textbf{Model analysis.} The agent reads the original model and tries to identify improvement opportunities, such as symmetry gaps (symmetric solutions can be eliminated by additional constraints), propagation weaknesses (nested loops can be replaced by global constraints), or representation opportunities (e.g., rewriting a 3D Boolean array as a 2D integer array).

\textbf{Model generation.} The agent generates a new, alternative model and saves it to a designated folder. Since the validation function treats all input models as interchangeable black boxes, the agent is instructed to ensure the alternative model satisfies three formatting requirements: (i) the whole model is built within a \texttt{build\_model()} function, which takes problem data as input and returns three things: the model, the decision variables, and the objective expression; (ii) it accepts the same input parameters as the original model; and (iii) it returns the same decision variables in the same order as the original model.

\textbf{Validation.} The agent invokes the validation function that evaluates the alternative model on the three training instances and compares it with the original model.

\textbf{Diagnosis and repair.} If the new model is incorrect, the agent creates a new model and re-validates it.

\textbf{Iteration.} Model generation, validation, and diagnosis and repair repeat until five \emph{correct} models have been generated. The agent is encouraged to create diverse model variants in each iteration. If twenty attempts have not produced five correct models, the agent should stop and report how far it got.

\textbf{Selection and logging.} The agent selects and saves the proposed model. Additionally, it stores a log file and a summary of the results. The log file reports explicitly which techniques the agent applies and why, and states the validation results. The agent is not given a fixed time limit, but the process is restricted to a maximum of 80 agent steps.

\section{Experimental Setup}
\label{sec:experiments}

\sloppypar
Supplementary material, including the prompt, validation function, original and proposed models, log files, and the training and test instances, is available on Zenodo~\footnote{\url{https://doi.org/10.5281/zenodo.22833583}}.

\subsection{Setup and Hardware}
All agent runs were executed on compute nodes with two Intel Xeon E5-2640 v4 processors (20 cores in total at 2.40\,GHz) and 157\,GiB RAM under Ubuntu~18.04; every individual run was capped at 8\,GiB of memory, and no GPUs were used. Python~3.13.9 invoked the agent and executed the CPMpy (version~1.0.0) models, which were solved with Chuffed~0.13.2 through MiniZinc~2.9.2.

We used the agentic Python coder~\cite{szeider_cp-agent_2026}, accessed through OpenRouter. As underlying model, we used nine different LLMs from six providers: Claude Fable~5, Claude Opus~5, and Claude Sonnet~5 from Anthropic, GPT~5.6 Sol and GPT~5.6 Terra from OpenAI, Gemini~3.7 Flash from Google, Grok~4.6 from xAI, as well as DeepSeek~4 Flash 0731 from DeepSeek, and Kimi K3 from Moonshot AI. All of them were run with a temperature of 0.3, except the two GPT~5.6 variants, which do not accept sampling parameters and were therefore used with their provider defaults.
 
The experiments reported here comprise 187 agent runs, which took a median of 7.0~minutes each and about 27~hours of agent wall-clock time in total. Within a validation, the performance on the training instances is compared in parallel on separate nodes. The ten test instances of an evaluation are handled in the same way. Thus, the reported run times of the agent include the parallel validation of the three training instances. Running the training instances sequentially would make a run correspondingly longer.

\subsection{Benchmark Problems and Instances}

We evaluate our approach on fifteen optimization problems encoded in CPMpy. Twelve of them are well-known combinatorial problems: Balanced Latin Rectangle%
, Job Shop Scheduling%
, Maximum Cut%
, Maximum Diversity%
, Maximum Weight Independent Set%
, Multi-dimensional Knapsack%
, Progressive Party%
, Quadratic Assignment%
, Resource-Constrained Project Scheduling%
, Set Cover%
, Traveling Salesperson%
, and Warehouse Location%
. The remaining three, namely \textit{Mentor Matching}, \textit{Cross-Dock Consolidation}, and \textit{Fleet Charging}, are new problems that we designed for this study. They are described below. We excluded symmetry-breaking constraints from all models, as these would limit the agent's room for improvement.

\begin{shaded*}
\textbf{Mentor Matching:} We are given a set of mentors with a minimum and maximum capacity, and a set of students. Mentors and students provide a list of topics of interest. The task is to assign students to mentors such that each mentor supervises a number of students within their capacity bounds. A mentor can only supervise a student if they share at least one topic of interest. The objective is to maximize the total number of shared interests across all assigned student--mentor pairs.
\end{shaded*}

Even though there exist many variants of matching problems, we are not aware of this specific formulation. Related problems, such as the Student--Project Allocation (SPA) problem~\cite{abraham_two_2007}, use preference lists rather than shared topics of interest.

\begin{shaded*}
\textbf{Cross-Dock Consolidation:} Packages arrive at a cross-dock and all of them have to be loaded onto outbound trucks on the same day. Every package has a weight and a destination zone. Every truck has a weight capacity, a set of zones it is allowed to serve, and a limit on the number of distinct zones it may visit. Every zone a truck visits incurs a stop fee, so packages for the same zone should be consolidated onto few trucks. Some pairs of packages are incompatible and must not travel on the same truck. The objective is to assign every package to a truck so that the total stop fees are minimal.
\end{shaded*}

The problem combines elements of bin packing with conflicts and of cross-docking, but we are not aware of this particular combination of a per-zone stop fee, a cap on the number of zones per truck, and pairwise package conflicts.

\begin{shaded*}
\textbf{Fleet Charging:} A depot charges its fleet of electric vehicles overnight, over a night divided into time slots. The depot owns chargers of several types; a charger of one type delivers a fixed amount of energy per slot, and the number of chargers of each type is given. Every vehicle is present during a time window, needs a given amount of energy, and can use only some of the charger types. A vehicle is plugged into one charger type for the whole night but may be charged in any subset of the slots of its window. A grid connection limits the total power drawn in every slot, and energy has a per-slot price. The objective is to choose a charger type and charging slots for every vehicle so that the total energy cost is minimal.
\end{shaded*}

Electric-vehicle charging schedules have been studied widely, but we are not aware of a formulation that combines charger types available in fixed numbers, interruptible charging within a vehicle's window, a per-slot site power cap, and time-of-use prices in this way.

We include novel problems for two reasons. First, they ensure that the agent cannot simply reproduce alternative models for the same problem from its training data. Second, they better represent our use case: the agent receives a newly created model for which no optimization has yet been considered.

For each problem, we generated three training and ten test instances. The original model must find at least one solution within 20 seconds on a training instance and within 60 seconds on a test instance. In both cases, we additionally required the solution found within the time limit not to be optimal, so that there remains room for improvement. For instances where optimality was uncertain, we ran the original model for 300 seconds (training) or 600 seconds (test) and verified that a strictly better solution was found within that extended time budget. Instances that did not satisfy these criteria were discarded and replaced. All benchmark problems and instances are available in the supplementary material.

\subsection{Solver-Side Variance}
\label{sec:variance}

To distinguish solver-side noise from real model-induced improvement, we ran each of the 150 test instances four times on the original model with different solver random seeds. For 102 instances, all four runs produced identical objective values. For 21 out of the remaining 48 instances, the relative deviation $(v_{\max}-v_{\min})/v_{\max}$ was less than 1.0\%. Individual comparisons can therefore fall either way. Repeating all 1500 test comparisons against each of the four original measurements in turn changes about 3\% of the verdicts and shifts the overall share of improved comparisons by less than two percentage points. The aggregate results therefore do not depend on which measurement is used.

\begin{table*}[t]
  \centering
  \caption{Comparison of nine LLMs on four problems, each with three training and ten test instances}
  \label{tab:llm_comparison}
  \newcommand{\zero}{\textcolor{gray!55}{0}}
  \begin{threeparttable}
  \begin{tabular}{@{}lrrr@{\hspace{10pt}}*{3}{w{r}{1.7em}}@{\hspace{10pt}}rrrrr@{}}
    \toprule
    & & & & \multicolumn{3}{c}{Training (of 12)} & \multicolumn{5}{c}{Test (of 40)} \\
    \cmidrule(lr){5-7}\cmidrule(l){8-12}
    Model & Cost (\$) & Time (min) & Steps & b & e & w & b & e & w & t & n \\
    \midrule
    Claude Fable~5    & 4.08 &  6.6 & 19.5 & 12 & \zero & \zero & 38 & \zero & 1 & \zero & 1  \\
    Claude Opus~5     & 5.25 & 14.9 & 32.0 & 12 & \zero & \zero & 35 & \zero & \zero & 4 & 1 \\
    GPT~5.6 Sol       & 0.22 &  6.6 & 29.5 & 9 & \zero & 3 & 33 & \zero & 7 & \zero & \zero \\
    Claude Sonnet~5   & 4.14 & 16.6 & 45.3 & 10 & 2 & \zero & 32 & 4 & 4 & \zero & \zero \\
    Grok~4.6          & 0.61 & 10.8 & 29.3 & 12 & \zero & \zero & 32 & \zero & 1 & \zero & 7  \\
    Gemini~3.7 Flash  & 0.19 &  5.8 & 28.8 & 10 & 2 & \zero & 31 & 4 & 5 & \zero & \zero \\
    DeepSeek~4 Flash  & 0.06 & 24.7 & 67.5 & 12 & \zero & \zero & 31 & \zero & 1 & 3 & 5 \\
    GPT~5.6 Terra     & 0.24 &  6.1 & 28.5 & 11 & \zero & 1 & 31 & \zero & 9 & \zero & \zero \\
    Kimi~K3$^\ast$    & 0.36 &  8.2 & 31.8 & 9 & \zero & \zero & 27 & \zero & 3 & \zero & \zero \\
    \bottomrule
  \end{tabular}
  \begin{tablenotes}[flushleft]\footnotesize
  \item[] The columns b, e, w, t and n count the outcomes \textit{better}, \textit{equal}, \textit{worse}, \textit{timeout}, and \textit{no\_solution}; the last two only occur on the test instances.
  \item[] $^\ast$Kimi~K3 produced no proposed model for Set Cover, so its outcomes cover three problems.
  \end{tablenotes}
  \end{threeparttable}
\end{table*}

\subsection{Evaluation Protocol}
\label{sec:evaluation_protocol}

Every run of our approach yields one proposed model, and only this model is considered for further evaluation. In the main experiment, we run our approach ten times for each of the fifteen benchmark problems; the comparison of LLMs in Section~\ref{sec:llm_comparison} uses the same evaluation on a smaller setup described there. To evaluate a proposed model, we execute the same method that is used in the agent's validation function as described in Section~\ref{validation}. We run the ten test instances with a timeout of 60 seconds. For the training instances, we use measurements reported by the agent.

We report whether the achieved objective value was better, equal, or worse than with the original model, or whether the proposed model is invalid. If it finds no solution within the time limit, we additionally run a feasibility check without the objective for 600 seconds, which separates a model that is merely too slow (\textit{timeout}) from one that finds no solution at all (\textit{no\_solution}). The solver's performance may vary across runs due to CPU load or other external artifacts; as the repeated measurements in Section~\ref{sec:variance} show, this changes about 3\% of the individual verdicts in the main experiment but shifts the overall share of improved comparisons by less than two percentage points. The evaluation follows a paired design: each (problem, instance) pair is solved by both the original and the proposed model under identical hardware, solver, and timeout, so the 1500 test comparisons of the main experiment control for instance-level difficulty.

\begin{table*}[t]
  \centering
  \caption{Results of the main experiment}
  \label{tab:main_results}
  \newcommand{\zero}{\textcolor{gray!55}{0}}
  \av{\footnotesize\setlength{\tabcolsep}{4pt}}%
  \begin{threeparttable}
  \begin{tabular}{@{}lrrr@{\hspace{14pt}}rrrrr@{\hspace{14pt}}rrr@{}}
    \toprule
    & \multicolumn{3}{c}{Training (of 30)} & \multicolumn{5}{c}{Test (of 100)} & \multicolumn{3}{c}{Test: relative objective change (\%)} \\
    \cmidrule(lr){2-4}\cmidrule(lr){5-9}\cmidrule(l){10-12}
    Problem & b & e & w & b & e & w & t & n & median gain & median loss & mean \\
    \midrule
    Quadratic Assignment            & 30 & \zero & \zero & 99 & \zero & 1 & \zero & \zero & $+6.1$ & $-0.7$ & $+6.6$ \\
    Mentor Matching$^\dagger$       & 30 & \zero & \zero & 98 & \zero & 1 & 1 & \zero & $+58.9$ & $-7.4$ & $+56.4$ \\
    Traveling Salesperson           & 30 & \zero & \zero & 92 & \zero & 8 & \zero & \zero & $+44.3$ & $-4.0$ & $+34.3$ \\
    Fleet Charging$^\dagger$        & 27 & \zero & 3 & 88 & 2 & 10 & \zero & \zero & $+62.0$ & $-1.2$ & $+47.3$ \\
    Maximum Diversity               & 30 & \zero & \zero & 88 & 2 & 10 & \zero & \zero & $+3.8$ & $-1.5$ & $+4.6$ \\
    Maximum Cut                     & 30 & \zero & \zero & 88 & 2 & 8 & \zero & 2 & $+13.0$ & $-1.3$ & $+14.3$ \\
    Balanced Latin Rectangle        & 28 & \zero & 2 & 86 & \zero & 14 & \zero & \zero & $+12.3$ & $-3.5$ & $+10.1$ \\
    Job Shop Scheduling             & 26 & \zero & 4 & 77 & \zero & 23 & \zero & \zero & $+9.9$ & $-11.0$ & $+12.6$ \\
    Progressive Party               & 29 & \zero & 1 & 73 & \zero & 27 & \zero & \zero & $+14.6$ & $-14.5$ & $+13.7$ \\
    Cross-Dock Consolidation$^\dagger$ & 28 & \zero & 2 & 71 & \zero & 25 & 4 & \zero & $+30.4$ & $-13.7$ & $+13.3$ \\
    RCPSP                           & 27 & 3 & \zero & 63 & 7 & 30 & \zero & \zero & $+2.5$ & $-3.8$ & $+0.9$ \\
    Set Cover                       & 18 & 7 & 5 & 38 & 10 & 52 & \zero & \zero & $+7.3$ & $-4.6$ & $+0.4$ \\
    Warehouse Location              & 14 & 6 & 10 & 35 & 19 & 46 & \zero & \zero & $+9.6$ & $-0.7$ & $+6.3$ \\
    Maximum Weight Independent Set  & 16 & 11 & 3 & 32 & 41 & 24 & \zero & 3 & $+1.6$ & $-1.4$ & $+0.2$ \\
    Multi-dimensional Knapsack      & \zero & 30 & \zero & \zero & 98 & 2 & \zero & \zero & --- & $-0.4$ & $-0.01$ \\
    \midrule
    All problems                    & 363 & 57 & 30 & 1028 & 181 & 281 & 5 & 5 & $+12.7$ & $-3.3$ & $+14.7$ \\
    \bottomrule
  \end{tabular}
  \begin{tablenotes}[flushleft]\footnotesize
  \item[] Fifteen problems, three training and ten test instances, ten runs per problem. The outcome columns are those of Table~\ref{tab:llm_comparison}. $^\dagger$~Problems we designed ourselves.
  \end{tablenotes}
  \end{threeparttable}
\end{table*}

\section{Results}
\label{sec:results}

Across the 150 runs of the main experiment, a run took a median of 30 steps: 19 Python executions, 10 updates of the agent's to-do list, and the one call that stores the selected model. The longest run took 43 steps, so the ceiling of 80 steps never came close to binding. A run consumed a median of 402K tokens and seven minutes of wall-clock time, and at most 1.8M tokens and 16 minutes. Reaching five validated candidate models took a median of six attempts and never more than eight, so candidates that turned out to be invalid or unsolvable cost at most three extra attempts.

\subsection{Comparison of Different LLMs}
\label{sec:llm_comparison}

In this experiment, we compared the nine LLMs listed in Section~\ref{sec:experiments} on four of the benchmark problems: Job Shop Scheduling, Quadratic Assignment, Set Cover, and Warehouse Location. Each model ran the agent once per problem on the problem's three training instances, and its proposed model was evaluated on the ten test instances as described in Section~\ref{sec:evaluation_protocol}. Table~\ref{tab:llm_comparison} summarizes the results. Cost, time, and steps are averages per problem; the cost is in US dollars, derived from the OpenRouter billing. Models are sorted by more \textit{better}, then more \textit{equal}, then fewer \textit{worse} test outcomes.

Kimi~K3 did not complete one of its runs. After validating a first model with the provided validation function, it went on to analyze the instances with self-written heuristics in Python, and then stopped without producing further models or selecting a proposed one. All other models produced a proposed model for every problem. These proposed models performed better than the original model on many of the training instances. Runs took about six minutes per problem on average for Claude Fable~5, GPT~5.6 Sol, Gemini~3.7~Flash, and GPT~5.6 Terra, and went up to 25 minutes for DeepSeek~4~Flash. The long runs coincide with many steps and repeated attempts to reach five correct models.

On the test instances, Claude Fable~5 achieved the most improvements with 38 of 40, followed by Claude Opus~5 with 35 and GPT~5.6 Sol with 33. However, the three Anthropic models, Claude Fable~5, Claude Opus~5, and Claude Sonnet~5, cost between 4.08 and 5.25 US dollars per problem, possibly also due to different caching behavior. Interestingly, Claude Opus~5 was the most expensive although Claude Fable~5 has the higher price per token. This was because Opus consumed about 2.5 times as many input tokens and about three times as many output tokens. Due to the high costs per run, we decided to use the cheaper GPT~5.6 Sol for the other experiments. Among the six inexpensive models, it achieved the most improvements, 33 of 40, within a short amount of time and without a single \textit{timeout} or \textit{no\_solution}. Claude Fable~5 achieved five more improvements, but at about 18 times the cost.

\subsection{Main Results}
\label{sec:main_results}

In the main experiment, we ran the agent 10 times on each of the 15 benchmark problems. We evaluated each proposed model on 10 held-back test instances, which gives 1500 paired comparisons. The proposed models reached a better objective value than the original in 1028 of these comparisons (68.5\%). Only ten comparisons returned no objective value at all. Every proposed model could be built and executed, so no comparison ended in a \textit{solver\_error}\footnote{In one of the 150 runs, the agent stored an unvalidated file as its proposed model. It could not be built, so all ten test comparisons ended in a \textit{solver\_error}. As this is detectable without any test instance, we discarded the run and repeated it.}. Table~\ref{tab:main_results} lists these outcomes per problem, sorted by the number of \textit{better} outcomes. Next to the test outcomes, it lists the outcomes on the three training instances, which the agent measured itself during the run and on which it selected the proposed model. The higher performance on the training instances is expected, since the agent selected the model that performed best on them.

The three columns on the right of Table~\ref{tab:main_results} quantify the test outcomes: median gain (resp.\ loss) is the median relative change of the objective value over the \textit{better} (resp.\ \textit{worse}) comparisons, and the mean is taken over all comparisons that returned an objective value. Over all comparisons, the mean relative improvement is $+14.7\%$, and a typical gain is about four times as large as a typical loss. Mentor Matching gains the most and Multi-dimensional Knapsack nothing at all. The reason for the weak performance of the multi-dimensional knapsack problem might be that the original model has a canonical formulation with one Boolean per item, one linear constraint per resource, and a linear objective. This leaves neither an expensive construct to replace nor a symmetry to break.

All runs stayed well inside the step budget, with 27 to 36 steps and six to ten and a half minutes on average per problem. Most importantly, not one of the 1500 comparisons was \textit{invalid}. Whenever a proposed model returned a solution, injecting that solution into the original model was satisfiable, so the solution was feasible for the original model and really achieved the reported objective value.

\begin{figure}[t]
  \centering
  \begin{tikzpicture}
    \begin{axis}[
      scale only axis, width=\dimexpr\columnwidth-2.95cm\relax, height=6.2cm,
      xlabel={mean relative improvement of a run (\%)},
      xmin=-28, xmax=68, xtick distance=20,
      ymin=0.4, ymax=15.6, ytick={1,2,3,4,5,6,7,8,9,10,11,12,13,14,15}, yticklabels={{Quadr. Assignment},{Mentor Matching},{Traveling Salesp.},{Fleet Charging},{Max. Diversity},{Max. Cut},{Bal. Latin Rect.},{Job Shop Sched.},{Progressive Party},{Cross-Dock Cons.},{RCPSP},{Set Cover},{Warehouse Loc.},{Max. Indep. Set},{Multi-Knapsack}},
      y dir=reverse,
      yticklabel style={font=\footnotesize},
      xticklabel style={font=\footnotesize}, xlabel style={font=\footnotesize},
      grid=major, major grid style={gray!25},
    ]
      \addplot[only marks, mark=*, mark size=1.1pt, color=black, opacity=0.45]
        coordinates {(4.3,1) (4.9,1) (5.4,1) (5.8,1) (6.4,1) (7.4,1) (7.7,1) (7.8,1) (8.1,1) (8.2,1) (50.6,2) (57.0,2) (57.1,2) (57.1,2) (57.1,2) (57.1,2) (57.1,2) (57.1,2) (57.1,2) (57.1,2) (0.4,3) (0.7,3) (39.0,3) (41.6,3) (41.7,3) (41.8,3) (41.9,3) (42.7,3) (43.3,3) (50.0,3) (0.1,4) (49.3,4) (51.2,4) (51.3,4) (53.0,4) (53.1,4) (53.2,4) (53.6,4) (53.7,4) (54.6,4) (3.0,5) (3.2,5) (3.2,5) (3.2,5) (3.3,5) (3.9,5) (3.9,5) (4.3,5) (4.3,5) (13.5,5) (12.0,6) (12.0,6) (12.3,6) (12.3,6) (13.2,6) (13.2,6) (13.8,6) (14.1,6) (18.7,6) (20.9,6) (7.7,7) (7.7,7) (8.4,7) (8.8,7) (9.5,7) (10.3,7) (10.5,7) (10.9,7) (13.3,7) (13.8,7) (-23.4,8) (-0.9,8) (4.0,8) (4.3,8) (4.5,8) (4.7,8) (4.7,8) (5.0,8) (61.5,8) (61.6,8) (4.8,9) (6.8,9) (7.3,9) (8.2,9) (11.5,9) (14.8,9) (18.7,9) (19.7,9) (22.3,9) (23.1,9) (1.7,10) (2.3,10) (2.5,10) (2.6,10) (5.8,10) (8.0,10) (23.9,10) (26.7,10) (33.8,10) (33.8,10) (-0.1,11) (-0.1,11) (0.6,11) (0.6,11) (0.9,11) (1.1,11) (1.1,11) (1.2,11) (1.3,11) (2.0,11) (-1.7,12) (-1.5,12) (-1.4,12) (-0.9,12) (-0.6,12) (-0.2,12) (0.5,12) (0.5,12) (0.8,12) (8.4,12) (-0.4,13) (-0.3,13) (-0.3,13) (-0.2,13) (8.0,13) (8.0,13) (8.0,13) (8.1,13) (8.1,13) (24.2,13) (-1.0,14) (-0.4,14) (-0.2,14) (0.1,14) (0.1,14) (0.2,14) (0.3,14) (0.6,14) (1.0,14) (1.3,14) (-0.1,15) (0.0,15) (0.0,15) (0.0,15) (0.0,15) (-0.0,15) (0.0,15) (0.0,15) (0.0,15) (0.0,15)};
      \addplot[only marks, mark=|, mark size=4pt, very thick, color=worse!70!black]
        coordinates {(6.6,1) (56.4,2) (34.3,3) (47.3,4) (4.6,5) (14.3,6) (10.1,7) (12.6,8) (13.7,9) (13.3,10) (0.9,11) (0.4,12) (6.3,13) (0.2,14) (-0.01,15)};
    \end{axis}
  \end{tikzpicture}
  \caption{Mean relative objective improvement of every individual run on its ten
  test instances, one dot per run, grouped by problem and ordered as in
  Table~\ref{tab:main_results}; the vertical bar marks the mean improvement of the problem given there.
  The problems differ not only in level but in spread: on Mentor Matching and
  Quadratic Assignment all ten runs land in a narrow band, whereas Job Shop
  Scheduling covers everything from $-23\%$ to $+62\%$.}
  \label{fig:run_spread}
\end{figure}
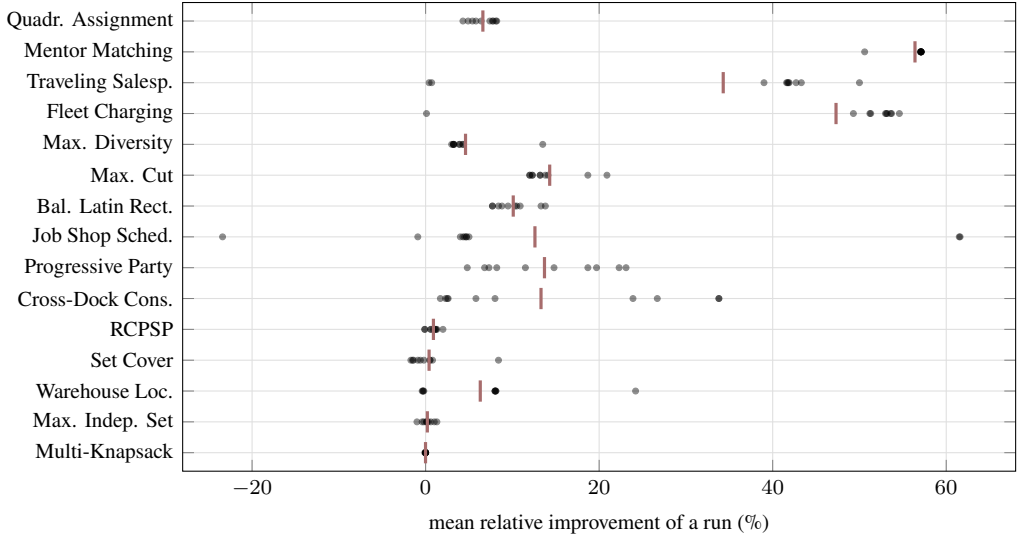

Figure~\ref{fig:run_spread} shows that the ten runs of a problem do not merely repeat each other. On Mentor Matching, Quadratic Assignment, and Maximum Diversity every run lands in a narrow band, whereas Job Shop Scheduling spreads over the whole range and Traveling Salesperson splits into a large group of strong runs and two weak ones. Where the spread is that wide, a single run says little about the problem, and repeated runs are needed to separate the approach from the variation between individual LLM runs. However, even though we only have three training instances, a strong performance on them seems to be a reliable indicator for a strong performance on the test instances. Runs that improved all three training instances improved 8.6 of the ten test instances on average, and runs that improved two, one, or none of them reached 4.9, 2.8 and 0.3 of ten. A weak run can therefore be discarded before any test instance is touched, which makes the three training instances useful not only for the agent's own search but also as a filter on its result.

\begin{table}[t]
  \centering
  \caption{Strategies applied by the proposed models}
  \label{tab:strategies}
  \newcommand{\zero}{\textcolor{gray!55}{0}}
  \setlength{\tabcolsep}{3.8pt}
  \begin{threeparttable}
  \begin{tabular}{@{}lrrrrrrr@{}}
    \toprule
    Problem & Glob. & View & Sym. & Impl. & Bnd. & Prep. & Ref. \\
    \midrule
    Quadratic Assignment        & 10 & 10 & 1 & 3 & \zero & 8 & \zero \\
    Mentor Matching$^\dagger$   & 10 & \zero & 5 & \zero & \zero & 10 & 10 \\
    Traveling Salesperson       & 8 & 8 & 10 & \zero & \zero & \zero & \zero \\
    Fleet Charging$^\dagger$    & 1 & 1 & 1 & 2 & \zero & 9 & 9 \\
    Maximum Diversity           & \zero & \zero & \zero & 7 & 9 & \zero & 9 \\
    Maximum Cut                 & \zero & \zero & 10 & 2 & 3 & \zero & 2 \\
    Balanced Latin Rectangle    & 10 & \zero & 10 & 3 & 5 & \zero & 10 \\
    Job Shop Scheduling         & 9 & \zero & 3 & 5 & 10 & \zero & 4 \\
    Progressive Party           & 10 & 2 & 10 & \zero & \zero & 1 & 5 \\
    Cross-Dock Consolid.$^\dagger$ & 4 & 4 & 4 & 8 & \zero & 10 & 9 \\
    RCPSP                       & 7 & 1 & 3 & 4 & 10 & 5 & \zero \\
    Set Cover                   & \zero & \zero & 2 & 2 & \zero & 10 & 1 \\
    Warehouse Location          & 1 & 1 & 1 & 5 & \zero & 1 & 8 \\
    Max.\ Weight Indep.\ Set    & \zero & \zero & \zero & 8 & 2 & \zero & 4 \\
    Multi-dim.\ Knapsack        & \zero & \zero & 1 & 3 & \zero & 3 & \zero \\
    \midrule
    All problems (of 150)       & 70 & 27 & 61 & 52 & 39 & 57 & 71 \\
    \bottomrule
  \end{tabular}
  \begin{tablenotes}[flushleft]\footnotesize
  \item[] Number of the ten runs per problem whose model applies the strategy.
  \end{tablenotes}
  \end{threeparttable}
\end{table}

\subsection{Applied Optimization Strategies}

When analyzing which optimization strategies were used by the agent, we could identify the following classifications: global constraint (Glob.), changed variable representation or viewpoint, often with channeling constraints linking auxiliary variables to the original output variables (View), symmetry breaking constraints (Sym.), implied constraints (Impl.), variable domains or objective bounds tightened with values computed in Python (Bnd.), preprocessing of the instance data such as fixing of variables or precomputed lookup tables (Prep.), and reformulated constraints or objectives within the same viewpoint (Ref.). Table~\ref{tab:strategies} shows which strategies were applied how often for which problem. The agent usually combined 2.5 strategies in one proposed model. Reformulations were used by 71 of the 150 models, followed by global constraints (70), symmetry breaking (61), and preprocessing (57).

Since manually categorizing all 150 models would have required substantial effort, we automated the categorization. Consequently, the table should be interpreted as an approximate overview of the optimization strategies used rather than as a definitive classification. The prompt used and the labels of every run are part of the supplementary material. 

\subsection{Non-Agentic Baselines}
\label{sec:singleshot}

To assess the contribution of the full agent protocol beyond simply sampling and selecting among several candidate models, we compare the agent against two non-agentic baselines that reuse the \emph{same} validation and selection tools but omit the agent's reasoning, iteration, and repair.
The \textbf{single-shot} baseline issues a single prompt to the same underlying LLM (GPT~5.6 Sol) with the same high-level modeling-technique guidance, asking for one improved CPMpy model, but without validation feedback, iteration, or error recovery. It has no fallback when its single candidate turns out to be invalid, which happens for four of the fifteen problems.
The \textbf{best-of-$k$} baseline generalizes this: $k$ independent single-shot candidates are generated, each is checked with the same validation procedure, and the best valid candidate (by training-instance performance) is selected. We report $k=5$, which matches the agent's \emph{generation} budget while withholding the agent's iterative feedback and repair: the agent is asked for five validated candidate models per run and needs a median of six attempts to obtain them. To keep the comparison on equal footing, the baselines and the agent-proposed models were evaluated on the same shared machine and software as all other reported results.

\begin{table}[t]
  \centering
  \caption{Non-agentic baselines vs.\ the agent}
  \label{tab:bestofk}
  \begin{threeparttable}
  \begin{tabular*}{\cv{\columnwidth}\av{0.62\columnwidth}}{@{\extracolsep{\fill}}lr@{}}
    \toprule
    Method & Better / 150 \\
    \midrule
    Single-shot ($k=1$) & 64\phantom{.8} \\
    Best-of-5 & 101\phantom{.8} \\
    \midrule
    Agent (ReAct), mean of ten runs & 102.8 \\
    \bottomrule
  \end{tabular*}
  \begin{tablenotes}[flushleft]\footnotesize
  \item[] Number of test instances (out of 150) on which the selected model beats the original, under a shared evaluation setup. Best-of-5 matches the number of candidate models the agent validates per run. The agent's value is the mean over its ten runs per problem.
  \end{tablenotes}
  \end{threeparttable}
\end{table}

Table~\ref{tab:bestofk} reports the comparison. Sampling several candidates and selecting among them helps considerably: moving from a single shot to best-of-5 raises the number of improved test instances from 64 to 101 of 150. With 101 improved instances, best-of-5 is on par with the agent, which improves 102.8 instances on average; drawing one agent run per problem gives a total with a standard deviation of 6.4. Larger samples do not change this picture: best-of-10 and best-of-20 improve 110 and 106 instances, respectively, both within about one standard deviation. With the same number of candidate models, the agent's iterative diagnosis and repair therefore yields no clear advantage in the aggregate improvement rate over independent sampling with the same validation and selection tools. This suggests that empirical validation and selection account for a large part of the gain. The agent retries until it has five valid models, while sampling only needs one valid candidate among its $k$. We expect the repair to pay off on models more complex than ours, where a valid candidate is rare enough that even twenty samples may contain none. An oracle that picks, for each problem, the one of twenty sampled candidates that is best on the \emph{test} instances improves 128 instances. It uses test data and is therefore an upper bound rather than a baseline, but its distance to best-of-20 shows that selection on three training instances often misses the best available candidate, which points to selection from few training instances as a limiting step.

\subsection{Anatomy of a Selected Run}
\label{sec:anatomy}

Here we describe one run of the agent in the main experiment on the Maximum Diversity problem using GPT~5.6 Sol. In only four of the 150 runs did the validation procedure reject an attempt as invalid. The considered run is one of them.

The Maximum Diversity problem asks to select exactly $p$ out of $n$ elements such that the sum of the pairwise distances between the selected elements is maximized. The original model uses one Boolean variable \texttt{select[i]} per element. The objective is stated as a sum of products of two such variables (Listing~\ref{list:md_original}). The agent used eight attempts to obtain five validated encodings. The run took 585 seconds and 36 steps. The agent documented each encoding in a docstring. In the following, text in \textit{italics} is quoted verbatim from these docstrings and from the agent's final summary.

\begin{lstlisting}[caption={Objective of the original model of the Maximum Diversity problem.},label=list:md_original,float]
model += (total_diversity == sum(
  distance[i][j] * select[i] * select[j]
  for i in range(n_items)
  for j in range(i + 1, n_items)))
\end{lstlisting}

\textbf{Attempts 1 and 2.} At first, the agent used a compact representation: an increasing sequence of $p$ integer variables that hold the indices of the selected elements. The pairwise distances were looked up in a flattened distance table, where the indices determine the required Boolean output variables (``\textit{Sorted selected-index view removes permutation symmetry.}''). However, this first attempt failed on all three training instances with a solver error, since it indexed a Python list with a CPMpy expression. After the agent wrapped the table in a CPMpy array, the approach worked on two training instances but no solution was found on the third. So this second attempt was also rejected. Consequently, the agent abandoned this representation.

\textbf{Attempt 3 (encoding 1).} Now the agent switched to a linearization of the objective. It introduced Boolean variables \texttt{both[k]} for all pairs $(i,j)$ of elements and tied them to the selection by the three inequalities $\mathit{both}_k\le\mathit{select}_i$, $\mathit{both}_k\le\mathit{select}_j$ and $\mathit{both}_k\ge\mathit{select}_i+\mathit{select}_j-1$. It replaced the products by a weighted sum of the pair variables. In the docstring, the agent notes that symmetry breaking is here not possible: ``\textit{Items are not interchangeable because their distance rows differ, so no sound item-order symmetry breaking exists.}'' This encoding worked on all three training instances.

\textbf{Attempt 4.} Then the agent tried to streamline the model by deleting the third inequality: ``\textit{Positive weights make the reverse direction implicit at every optimum.}'' However, the validation procedure rejected this attempt as being invalid on two training instances. On the third instance, it performed worse than the original. The validator reports only the verdict, not the reason. However, the reason can still be worked out since this attempt differs from encoding 1 only by the deleted inequality. The agent worked out that the objective adds the distance of a pair only if its pair variable is 1 and all distances are positive. Consequently, if both elements of a pair are selected, an optimal solution sets the pair variable to 1 in any case and hence the deleted inequality is not needed. This holds for optimal solutions. Under the time limit, however, the solver returns the best solution found so far, in which a pair variable can be 0, even though both of its elements are selected. The reported objective value is then lower than the diversity of the selection. The original model, into which the validation procedure injects both the selection and the objective value, has no solution.

\textbf{Attempt 5 (encoding 2).} The agent restored an exact link by replacing the three inequalities with the reified conjunction \texttt{both[k] == (select[i] \& select[j])} (``\textit{Pair Booleans are equivalently channeled with reified conjunctions, replacing quadratic products by a weighted linear objective.}''). This encoding was valid but performed worse than the original on one training instance.

\textbf{Attempts 6 to 8 (encodings 3 to 5).} In encoding 3, the agent kept all three inequalities but added the implied constraint that exactly $p(p-1)/2$ pair variables are true. In encoding 4, it returned to its refuted idea but repaired it to be sound. With the third inequality not being present, the implied pair count forces the pair variable of every selected pair to be true (``\textit{Together these enforce equivalence with fewer primitive inequalities.}''). This encoding was valid but again performed worse than the original on one training instance. Encoding 5 combines an exact product definition of the pair variables with the implied pair count (Listing~\ref{list:md_improved}). Now encodings 1, 3 and 5 were all better than the original on all three training instances. The agent selected encoding 5, which achieved the best value of the run on the third training instance.

\begin{lstlisting}[caption={Pair variables and objective of the proposed model of the Maximum Diversity problem.},label=list:md_improved,float]
for k, (i, j) in enumerate(pairs):
  model += both[k] == select[i] * select[j]
model += sum(both) == n_select * (n_select - 1) // 2
model += total_diversity == sum(
  distance[i][j] * both[k]
  for k, (i, j) in enumerate(pairs))
\end{lstlisting}

The proposed model was indeed better than the original model on all ten test instances with relative gains between 0.03\% and 10.4\%. The agent summarized the encoding as follows: ``\textit{The final formulation explicitly linearizes selected pairs and adds the implied exact pair-cardinality constraint.}'' The objective is linear in the pair variables and the product in Listing~\ref{list:md_improved} defines each pair variable in terms of two Boolean variables. The run exemplifies why the agent's correctness arguments need to be checked by the validator. The argument behind attempt 4 was plausible, but injecting its solutions into the original model exposed the attempt as invalid.

\section{Discussion and Conclusion}
\label{sec:conclusion}

We introduced a novel agentic framework that automatically improves constraint models. It takes a baseline CPMpy model and only three small training instances as input, and iteratively creates and tests alternative model versions, applying various optimization strategies. In the end, the model considered best-performing is selected and returned. We tested our approach on a subset of four problems to evaluate the performance on different LLMs. Even though the models proposed by Claude Fable~5 improved most of the test instances (38 of 40), we decided to use GPT~5.6 Sol for the remaining experiments, since it showed a solid improvement rate (33 of 40) at only one-eighteenth of the price of Fable.
Across fifteen benchmark problems with ten test instances each and ten agent runs, our approach performed better in 1028 of 1500 cases (68.5\%) and yields a mean relative objective change of 14.7\%. The results suggest that open-ended, empirically validated reformulation is a practical alternative to fixed-rule reformulation systems, and that LLM agents can design improved constraint models even with only a very small number of training instances. A best-of-5 sampling baseline that reuses the same validation and selection tools reaches a comparable improvement rate (101 of 150 test instances, against 102.8 on average for the agent), so with this number of candidate models the aggregate gain does not depend on the agent's iterative diagnosis and repair.

In our experiments, the agent applied a diverse set of model optimization strategies, including reformulations, global constraints, symmetry breaking,
and preprocessing. It even combined several of these strategies, suggesting that LLM-based agents can reason about modeling decisions beyond simple template-based modifications.

A few aspects suggest natural follow-up work. Semantic correctness is checked empirically, by injecting candidate solutions into the original model, rather than proved; this was reliable in our experiments, where no proposed model was invalid on a test instance, but a formal equivalence or optimality-preservation guarantee would be a useful addition. The approach also benefits from solvers that make gradual progress on the training instances, so that performance differences are observable, and is less informative when improvements are rare. Finally, we removed existing symmetry-breaking constraints to leave room for improvement; quantifying the gains on already well-engineered models is a natural next step.

We see several potential directions for future work. A first natural next step could be to extend our approach to constraint satisfaction problems, where, instead of improvements to the objective value, speedups in running time could be used as an indicator of good models. Additionally, the agent could generate the training instances itself, so the user does not need to provide them. Another interesting direction would be to apply our approach in a real-time scenario with only a single optimization instance. In this setting, the original model could run continuously, while the agent simultaneously searches for improved models by evaluating them on the same instance with shorter time limits. Once a model has been proposed, it can be used for the remaining solving time. Future work could also include further ablation studies on individual components of the framework, like different numbers of training instances or asking the agent to produce a different number of models before deciding for one.

\section*{Acknowledgements}

Research was supported by the Austrian Science Fund (FWF) within the projects 10.55776/COE12 and 10.55776/P36420.

This work was presented at LLM-Solve 2026, a FLoC \& CP 2026 workshop, on July 19, 2026.

The authors used Claude Code with the models Opus and Fable for parts of the implementation, the search of established benchmark problems, the writing of the baseline CPMpy models, the generation of the instances, the categorization reported in Table~\ref{tab:strategies}, and the drafting of parts of the text. The authors directed this work, reviewed and tested all of it, and take full responsibility for the content.

\bibliographystyle{IEEEtran}
\bibliography{bibliography}

\begin{thebibliography}{10}
\providecommand{\url}[1]{#1}
\csname url@samestyle\endcsname
\providecommand{\newblock}{\relax}
\providecommand{\bibinfo}[2]{#2}
\providecommand{\BIBentrySTDinterwordspacing}{\spaceskip=0pt\relax}
\providecommand{\BIBentryALTinterwordstretchfactor}{4}
\providecommand{\BIBentryALTinterwordspacing}{\spaceskip=\fontdimen2\font plus
\BIBentryALTinterwordstretchfactor\fontdimen3\font minus
  \fontdimen4\font\relax}
\providecommand{\BIBforeignlanguage}[2]{{%
\expandafter\ifx\csname l@#1\endcsname\relax
\typeout{** WARNING: IEEEtran.bst: No hyphenation pattern has been}%
\typeout{** loaded for the language `#1'. Using the pattern for}%
\typeout{** the default language instead.}%
\else
\language=\csname l@#1\endcsname
\fi
#2}}
\providecommand{\BIBdecl}{\relax}
\BIBdecl

\bibitem{handbook_symmetry}
I.~P. Gent, K.~E. Petrie, and J.~Puget, ``Symmetry in constraint programming,''
  in \emph{Handbook of Constraint Programming}, ser. Foundations of Artificial
  Intelligence.\hskip 1em plus 0.5em minus 0.4em\relax Elsevier, 2006, pp.
  329--376.

\bibitem{handbook_implied}
B.~M. Smith, ``Modelling,'' in \emph{Handbook of Constraint Programming}, ser.
  Foundations of Artificial Intelligence.\hskip 1em plus 0.5em minus
  0.4em\relax Elsevier, 2006, pp. 377--406.

\bibitem{handbook_global_constraints}
W.~van Hoeve and I.~Katriel, ``Global constraints,'' in \emph{Handbook of
  Constraint Programming}, ser. Foundations of Artificial Intelligence.\hskip
  1em plus 0.5em minus 0.4em\relax Elsevier, 2006, pp. 169--208.

\bibitem{cp_handbook_2006}
F.~Rossi, P.~van Beek, and T.~Walsh, Eds.,
  \emph{\BIBforeignlanguage{en}{Handbook of {Constraint} {Programming}}}, ser.
  Foundations of {Artificial} {Intelligence}.\hskip 1em plus 0.5em minus
  0.4em\relax Elsevier, 2006, vol.~2.

\bibitem{akgun_conjure_2022}
{\"O}.~Akg{\"u}n, A.~M. Frisch, I.~P. Gent, C.~Jefferson, I.~Miguel, and
  P.~Nightingale, ``{Conjure}: Automatic generation of constraint models from
  problem specifications,'' \emph{Artificial Intelligence}, vol. 310, p.
  103751, Sep. 2022.

\bibitem{nightingale_savilerow_2017}
P.~Nightingale, {\"O}.~Akg{\"u}n, I.~P. Gent, C.~Jefferson, I.~Miguel, and
  P.~Spracklen, ``Automatically improving constraint models in {Savile}
  {Row},'' \emph{Artificial Intelligence}, vol. 251, pp. 35--61, Oct. 2017.

\bibitem{automatic_symmetry_detection}
S.~J. Zhang, J.~Sun, C.~Sun, Y.~Liu, J.~Ma, and J.~S. Dong, ``Constraint-based
  automatic symmetry detection,'' in \emph{2013 28th IEEE/ACM International
  Conference on Automated Software Engineering (ASE)}, 2013, pp. 15--25.

\bibitem{spracklen_automated_2023}
P.~Spracklen, N.~Dang, {\"O}.~Akg{\"u}n, and I.~Miguel, ``Automated streamliner
  portfolios for constraint satisfaction problems,'' \emph{Artificial
  Intelligence}, vol. 319, p. 103915, Jun. 2023.

\bibitem{satzilla_2008}
L.~Xu, F.~Hutter, H.~H. Hoos, and K.~Leyton-Brown, ``{SATzilla}:
  portfolio-based algorithm selection for {SAT},'' \emph{Journal of Artificial
  Intelligence Research}, vol.~32, no.~1, pp. 565--606, Jun. 2008.

\bibitem{lindauer_autofolio_2015}
M.~Lindauer, H.~H. Hoos, F.~Hutter, and T.~Schaub,
  ``\BIBforeignlanguage{en}{{AutoFolio}: An automatically configured algorithm
  selector},'' \emph{\BIBforeignlanguage{en}{Journal of Artificial Intelligence
  Research}}, vol.~53, pp. 745--778, Aug. 2015.

\bibitem{hutter_smac_2011}
F.~Hutter, H.~H. Hoos, and K.~Leyton-Brown,
  ``\BIBforeignlanguage{en}{Sequential model-based optimization for general
  algorithm configuration},'' in \emph{\BIBforeignlanguage{en}{Learning and
  {Intelligent} {Optimization}}}.\hskip 1em plus 0.5em minus 0.4em\relax
  Berlin, Heidelberg: Springer, 2011, pp. 507--523.

\bibitem{mechqrane_active_2026}
Y.~Mechqrane and C.~Bessiere, ``\BIBforeignlanguage{en}{Active constraint
  acquisition using large language models},''
  \emph{\BIBforeignlanguage{en}{Journal of Artificial Intelligence Research}},
  vol.~85, Feb. 2026.

\bibitem{regin_combining_2024}
F.~Régin, E.~De~Maria, and A.~Bonlarron, ``Combining constraint programming
  reasoning with large language model predictions,'' in \emph{30th
  {International} {Conference} on {Principles} and {Practice} of {Constraint}
  {Programming} ({CP} 2024)}, ser. {LIPIcs}, vol. 307, 2024, pp. 25:1--25:18.

\bibitem{michailidis_constraint_2024}
K.~Michailidis, D.~Tsouros, and T.~Guns, ``Constraint modelling with {LLMs}
  using in-context learning,'' in \emph{30th {International} {Conference} on
  {Principles} and {Practice} of {Constraint} {Programming} ({CP} 2024)}, ser.
  {LIPIcs}, vol. 307, 2024, pp. 20:1--20:27.

\bibitem{DBLP:conf/ecai/MichailidisTG25}
------, ``{CP}-{Bench}: Evaluating large language models for constraint
  modelling,'' in \emph{28th European Conference on Artificial Intelligence
  ({ECAI} 2025)}, ser. Frontiers in Artificial Intelligence and Applications,
  vol. 413.\hskip 1em plus 0.5em minus 0.4em\relax {IOS} Press, 2025, pp.
  861--868.

\bibitem{song_llms_2026}
Y.~Song and E.~Cohen, ``\BIBforeignlanguage{en}{Do {LLMs} understand constraint
  programming? {Zero}-shot constraint programming model generation using
  {LLMs}},'' in \emph{\BIBforeignlanguage{en}{Learning and {Intelligent}
  {Optimization}}}.\hskip 1em plus 0.5em minus 0.4em\relax Cham: Springer
  Nature Switzerland, 2026, pp. 16--31.

\bibitem{szeider_cp-agent_2026}
S.~Szeider, ``{CP-Agent}: Agentic constraint programming,'' in
  \emph{Proceedings of the 3rd International Workshop on Large Language Models
  For Code ({LLM4Code} 2026)}.\hskip 1em plus 0.5em minus 0.4em\relax {ACM},
  2026, pp. 6--13,
  doi:\,\href{https://doi.org/10.1145/3786181.3788711}{10.1145/3786181.3788711}.

\bibitem{pellegrino_when_2025}
A.~Pellegrino and J.~Mauro, ``When words change the model: Sensitivity of
  {LLMs} for constraint programming modelling,'' Nov. 2025, arXiv:2511.14334
  [cs].

\bibitem{voboril_generating_2025}
F.~Voboril, V.~Peruvemba~Ramaswamy, and S.~Szeider,
  ``\BIBforeignlanguage{en}{Generating streamlining constraints with large
  language models},'' \emph{\BIBforeignlanguage{en}{Journal of Artificial
  Intelligence Research}}, vol.~84, Oct. 2025,
  doi:\,\href{https://doi.org/10.1613/jair.1.18965}{10.1613/jair.1.18965}.

\bibitem{voboril_streamllm_2025}
------, ``\BIBforeignlanguage{en}{{StreamLLM}: Enhancing constraint programming
  with large language model-generated streamliners},'' in
  \emph{\BIBforeignlanguage{en}{2025 {IEEE}/{ACM} 1st {International}
  {Workshop} on {Neuro}-{Symbolic} {Software} {Engineering} ({NSE})}}.\hskip
  1em plus 0.5em minus 0.4em\relax Ottawa, ON, Canada: IEEE, May 2025, pp.
  17--22,
  doi:\,\href{https://doi.org/10.1109/NSE66660.2025.00010}{10.1109/NSE66660.2025.00010}.

\bibitem{voboril_balancing_2025}
------, ``Balancing {Latin} rectangles with {LLM}-generated streamliners,'' in
  \emph{31st {International} {Conference} on {Principles} and {Practice} of
  {Constraint} {Programming} ({CP} 2025)}, ser. {LIPIcs}, vol. 340, 2025, pp.
  36:1--36:17,
  doi:\,\href{https://doi.org/10.4230/LIPIcs.CP.2025.36}{10.4230/LIPIcs.CP.2025.36}.

\bibitem{cpmpy}
\BIBentryALTinterwordspacing
T.~Guns, ``Increasing modeling language convenience with a universal
  n-dimensional array, {CPpy} as python-embedded example,'' in
  \emph{Proceedings of the 18th International Workshop on Constraint Modelling
  and Reformulation ({ModRef} 2019)}, 2019, pp. 1--8. [Online]. Available:
  \url{https://archive.modref.org/files/papers/2019/ModRef2019-04-Increasing-modeling-language-convenience-with-a-universal-n.pdf}
\BIBentrySTDinterwordspacing

\bibitem{yao_react_2022}
\BIBentryALTinterwordspacing
S.~Yao, J.~Zhao, D.~Yu, N.~Du, I.~Shafran, K.~R. Narasimhan, and Y.~Cao,
  ``\BIBforeignlanguage{en}{{ReAct}: Synergizing reasoning and acting in
  language models},'' in \emph{\BIBforeignlanguage{en}{The {Eleventh}
  {International} {Conference} on {Learning} {Representations} ({ICLR} 2023)}},
  2023. [Online]. Available: \url{https://openreview.net/forum?id=WE_vluYUL-X}
\BIBentrySTDinterwordspacing

\bibitem{abraham_two_2007}
D.~J. Abraham, R.~W. Irving, and D.~F. Manlove, ``\BIBforeignlanguage{en}{Two
  algorithms for the {Student}-{Project} {Allocation} problem},''
  \emph{\BIBforeignlanguage{en}{Journal of Discrete Algorithms}}, vol.~5,
  no.~1, pp. 73--90, Mar. 2007.

\end{thebibliography}

\end{document}